\pdfoutput=1

\documentclass[11pt]{article}

\usepackage{acl2023}

\usepackage{times}
\usepackage{latexsym}
\usepackage{url}
\usepackage{graphicx}
\usepackage{booktabs}
\usepackage{caption}

\usepackage[T1]{fontenc}

\usepackage[utf8]{inputenc}

\usepackage{microtype}

\usepackage{inconsolata}

\usepackage{textgreek} 

\usepackage{graphicx}

\title{\textsc{CebuaNER}: A New Baseline Cebuano Named Entity Recognition Model}

\author{Ma. Beatrice Emanuela Pilar$^{\Omega}$~\;~Ellyza Mari Papas$^{\Omega}$~\;~Mary Loise Buenaventura$^{\Omega}$\\ \textbf{Dane Dedoroy$^{\Omega}$~\;~Myron Darrel Montefalcon$^{\Lambda}$~\;~Jay Rhald Padilla$^{\Lambda}$~\;~Lany Maceda$^{\Sigma}$} \\ \textbf{Mideth Abisado}$^{\Lambda}$ \and \textbf{Joseph Marvin Imperial}$^{\Lambda,\Gamma}$ \\
$^{\Omega}$Silliman University, Philippines $^{\Sigma}$Bicol University, Philippines \\ $^{\Lambda}$National University, Philippines, $^{\Gamma}$University of Bath, UK\\
\texttt{\href{mailto:beatricenpilar@su.edu.ph}{beatricenpilar@su.edu.ph},\href{mailto:jrimperial@national-u.edu.ph}{jrimperial@national-u.edu.ph}}
}

\begin{document}
\maketitle
\begin{abstract}

Despite being one of the most linguistically diverse groups of countries, computational linguistics and language processing research in Southeast Asia has struggled to match the level of countries from the Global North. Thus, initiatives such as open-sourcing corpora and the development of baseline models for basic language processing tasks are important stepping stones to encourage the growth of research efforts in the field. To answer this call, we introduce \textsc{CebuaNER}, a new baseline model for named entity recognition (NER) in the Cebuano language. Cebuano is the second most-used native language in the Philippines with over 20 million speakers. To build the model, we collected and annotated over 4,000 news articles, the largest of any work in the language, retrieved from online local Cebuano platforms to train algorithms such as Conditional Random Field and Bidirectional LSTM. Our findings show promising results as a new baseline model, achieving over 70\% performance on precision, recall, and F1 across all entity tags as well as potential efficacy in a crosslingual setup with Tagalog.
\end{abstract}

\section{Introduction}
Open-sourced and accessible machine-readable language datasets drive the progress of computational linguistics research. As such, university and industry research initiatives such as IndoNLP \cite{wilie-etal-2020-indonlu,aji-etal-2022-one}, Glot500 \cite{imanigooghari-etal-2023-glot500}, MasakhaneNER \cite{adelani-etal-2021-masakhaner,adelani-etal-2022-masakhaner} as well as conferences like Language Resources and Evaluation (LREC)\footnote{\url{http://www.lrec-conf.org/}} encourage and advocate for increased efforts in developing and release of high-quality resources to the community. Despite these efforts, however, languages in other parts of the world, such as in South East Asian (SEA) countries like the Philippines, Thailand, and Myanmar, still remain on the lower end of the level of digital support by researchers \cite{simons-etal-2022-assessing}. 

In Natural Language Processing (NLP) research, Named Entity Recognition (NER) is the task of labeling identifiable entities such as organization name (\textit{"Tottenham Hotspurs"}, \textit{"Red Cross"}) and specific locations (\textit{"Manila City"}, \textit{"Penny Lane Street"}) as in texts. It is considered one of the foundational information extraction tasks in NLP that are used frequently by both the research community and the industry \cite{loricanathan,vajjala-balasubramaniam-2022-really}. A good NER model serves as a backbone for more advanced systems requiring a deeper understanding of contextual semantics and disambiguation of texts to retrieve insights \cite{zhou-etal-2019-dual}. To date, research on NER has focused on improving the performances of models through advanced methods. Architectural additions such as predefined entity lists like gazetteers \cite{rijhwani-etal-2020-soft}, data augmentation techniques \cite{yaseen-langer-2021-data,cai-etal-2023-graph}, and complex neural methods \cite{chiu-nichols-2016-named,cotterell-duh-2017-low,liu2018empower,zhou-etal-2019-dual} have been used. Likewise, a plethora of online tools such as \textsc{spaCy}\footnote{\url{https://spacy.io/models/xx}} and \textsc{Stanza}\footnote{\url{https://stanfordnlp.github.io/stanza/ner\_models.html}} already integrates production-ready NER models for high-resource languages such as English, Chinese, and German.

In this study, we introduce \textbf{\textsc{CebuaNER}}, a new baseline named entity recognition model for the language Cebuano as a response to the call for new initiatives of tool, model, and dataset creation for low-resource languages. We collected and annotated over 4,000 articles written in Cebuano to train NER models using modern machine learning algorithms such as including Conditional Random Fields (CRF) and Bidirectional Long Short-Term Memory (Bi-LSTM). We specifically selected the task of NER for our study's contribution because of its simplicity and potential to serve as a baseline resource for advanced initiatives in computational linguistics and NLP for the Cebuano language. NER extracts essential information from unstructured texts by identifying and classifying named entities, making it easier for computational analysis to be more meaningful and context-sensitive \cite{pant2023named}. It also helps organize and categorize language data, providing valuable insights into language patterns and usage. This is particularly important for languages with limited digital resources. Additionally, NER is instrumental in creating digital dictionaries and grammar tools essential for academic understanding and language learning. These resources make languages more accessible and user-friendly for current and future research initiatives in Cebuano \cite{gharagozlou2023semantic}. From this paper, we hope to inspire more efforts to develop and improve the digital representation of Cebuano and other under-resourced Philippine languages through open sourcing and making our code and data publicly available\footnote{\url{https://github.com/mebzmoren/CebuaNER}}.


\section{Previous Works}

In the past years, studies in named entity recognition (NER) for Philippine languages have mainly focused on Filipino due to the ease of access to raw data. One of the first few works to use machine learning-based modeling is the study of \citet{alfonso2013named} using Conditional Random Fields on a dataset of biographies. The model was able to detect standard text entities such as people, organization, and location at a performance measure of 83\% in F1 score. The study reported difficulty with discriminating places and organizations with 42\% and 33\% error rates, respectively. A following study by \citet{ebona2013named} was published using Maximum Entropy on a Filipino short story dataset with a performance 80.53\% in F1 score. Similar to \citet{alfonso2013named}, the model also struggled in identifying location and organization information with error rates of 29.41\% and 13.10\%, respectively. More recently, the work of \citet{cruz2018named} also used Conditional Random Fields but on a compiled news article dataset achieving 75.71\% overall F1 score.

Aside from works on Filipino data, there are small research efforts to adapt the NER methodology for the Cebuano language. However, most of these works claim to be preliminary results due to the limited availability of gold-standard annotations. The work of \cite{maynard2003ne} first attempted to adapt an English NER system called ANNIE to Cebuano. The study involved replacing modules of tokenization, lexicon, and gazetteers from a small annotated Cebuano news dataset. The system achieved a promising performance of 69.1\% in F1 score, reporting possible sources of error in untrained human annotators for the named entity recognition task. Upon checking, the Cebuano NER module in ANNIE is not publicly available. A subsequent study by \citet{cotterell-duh-2017-low} examined a trained neural CRF on Filipino in a crosslingual setup using a separate silver-standard Cebuano data from Wikipedia. The neural CRF's performance was slightly lower than the log-linear CRF on Tagalog alone (56.98\% vs. 58.15\%). Nevertheless, when incorporating cross-lingual data from Cebuano, the neural CRF demonstrated significant improvement, outperforming the log-linear CRF by achieving an F1 score of 81.79\% compared to 75.29\%. More recently, a study by \citet{gonzales2022developing} proposed a hybrid neural network method for both part-of-speech tagging and NER. The work reported preliminary results with approximately 95-98\% in both precision and recall but only used a small dataset of 200 news articles.


Our study's major difference from these preliminary efforts is that we start from the ground up in terms of training NER models. We build a large gold-standard Cebuano dataset composed of 4,258 new articles annotated with high reliability by native speakers, which will be made open-sourced upon publication. Our dataset was sourced from recent content published by local Cebuano news platforms within the last five years. We see this as another advantage of this work, as recency and being able to capture modern language changes is an important aspect of automated tools. Lastly, compared to other works mentioned, we explore and compare the performances of modern machine learning algorithms for baseline model development which have shown greater effectivity for the task, especially for low-resource languages \cite{cotterell-duh-2017-low,zhou-etal-2019-dual}.


\begin{figure}[!htbp]
    \centering
    \includegraphics[width=.60\textwidth,trim={5.5cm 5.0cm 2.3cm 1cm}, clip]{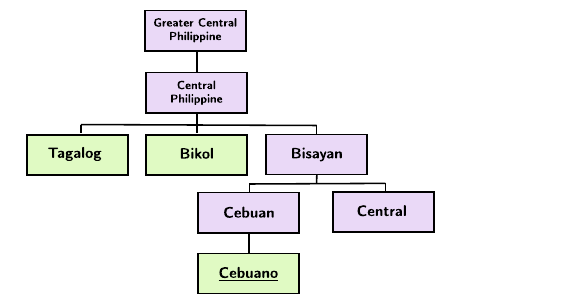}
    \caption{The central subgroup of the Philippine language family tree highlighting the origin of Cebuano language (\textsc{CEB}). Adapted with permission from \citet{imperial-etal-2022-baseline}.}
    \label{fig:cebuano_tree}
\end{figure}

\section{The Cebuano Language (\textsc{CEB})}
The Philippines is one of the most linguistically diverse countries in Asia \cite{mcfarland2008linguistic,metila2016challenge}. Part of the nation's linguistic identity is Cebuano (\textsc{CEB})\footnote{\url{https://www.ethnologue.com/language/ceb/}} which is the second most widely spoken language with over 27 million active speakers next to the national language Filipino. As part of the Bisayan language family, Cebuano exhibits a vibrant linguistic heritage and is spoken in the regions of Cebu, Siuijor, Bohol, Negros Oriental, northeastern Negros Occidental, southern Masbate, and in central areas of Mindanao. Despite this considerable number of speakers, Cebuano still continues to be classified as an under-resourced language by most data survey papers due to its very limited digital support \cite{imperial-etal-2022-baseline,simons-etal-2022-assessing}. We illustrate the placement of the Cebuano language in the Greater Central Philippine family tree in Figure~\ref{fig:cebuano_tree}.

\section{Corpus Building and Preprocessing}
This section of the paper presents a comprehensive outline of our procedure for building a Cebuano corpus. The following steps are taken to accomplish this task: data collection, annotation, and reliability testing.

\subsection{Data Collection}
For collecting Cebuano data, we collected publicly available articles from two local news sources in Cebuano, \textbf{Yes the Best Dumaguete} and the \textbf{Filipinas Bisaya}. To further increase the data count, we also incorporated another publicly available dataset from \textbf{SunStar Cebu} pre-collected by independent researcher Arjemariel Requina\footnote{\url{https://github.com/rjrequina/Cebuano-POS-Tagger}}. The total accumulated and filtered size of the Cebuano dataset is 4,258 articles. Table~\ref{tab:data} presents the distribution of the dataset per source. 

\begin{table}[!htbp]
\centering
\small
\begin{tabular}{lrr} 
 \toprule
 \textbf{Source} & \textbf{Original} & \textbf{Cleaned} \\ 
  \midrule
 Yes the Best Dumaguete & 1,484 & 781 \\ 
 Filipinas Bisaya & 769 & 377 \\ 
 SunStar Cebu & 3,100 & 3,100 \\
 \bottomrule
\end{tabular}
\caption{Statistics of news data sources for building \textsc{CebuaNER}.}
\label{tab:data}
\end{table}

\subsection{Annotation Process} 
\begin{figure}[!htpb]
    \centering
    \includegraphics[width=1\linewidth]{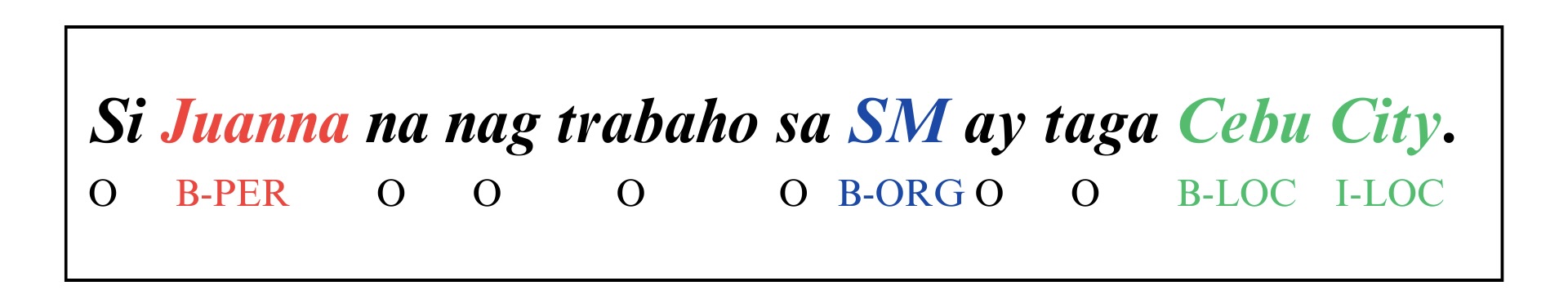}
    \caption{Cebuano sentence with annotations}
    \label{fig:sample_annotation}
\end{figure}

In the annotation process of the Cebuano dataset, we used Label Studio, an open-sourced data labeling platform\footnote{\url{https://labelstud.io/}}.  We employed and trained two undergraduate students who are native speakers of the Cebuano language for the labeling task. To follow labeling formats of current research in NER \cite{mayhew-roth-2018-talen,mayhew-etal-2019-named,adelani-etal-2021-masakhaner}, we annotated four entity types through the BIO encoding schema and used the tags Person (PER), Organization (ORG), Location (LOC), and Other (OTHER). We show an example of how a text in Cebuano is annotated using these tags in Figure~\ref{fig:sample_annotation}. 

\subsection{Reliability Testing}
We noticed that previous works mentioned in Section 2, especially for NER in Philippine languages, lack information about how reliable the tags in their respective datasets are. We see this as a limitation that should be avoided as transparency of data quality is important for progress in the field. Thus, for this study, we calculate the reliability of annotations of the tags in our annotated Cebuano dataset. We use Cohen's $\kappa$ \cite{cohen1960coefficient} as done in previous works for NER such as in \citet{balasuriya2009named,brandsen2020creating,jarrar-etal-2022-wojood}. Cohen's $\kappa$ involves comparing the observed agreement $p_{o}$ between annotators to the agreement that would be expected by chance $p_{e}$ using the formula:
\begin{equation}
    \kappa = \frac{p_{o}-p_{e}}{1-p_{e}}
\end{equation}


Table~\ref{tab:cohens_kappa} shows the agreement scores between annotators. The observed agreement indicates that around 98.37\% of the data points have labels on which the annotators agree, demonstrating a high level of consistency in their annotations. The agreement by chance represents the proportion of agreement that would be expected by random chance alone. As it is lower than the observed agreement, it suggests that the annotators' agreement exceeds what would be expected by chance. A Cohen's $\kappa$ score that is close to 1.0 implies a high level of agreement. Thus, a value of 0.9315 obtained in our study further supports the notion of strong agreement between the annotators.

\begin{table}[!htbp]
\small
\centering
\begin{tabular}{lr} 
 \toprule
 Observed Agreement & 0.9837 \\ 
 Agreement by Chance & 0.7617 \\ 
 \textbf{Cohen's $\kappa$} & \textbf{0.9315} \\
 \bottomrule
\end{tabular}
\caption{Cohen's $\kappa$ results from annotations.}
\label{tab:cohens_kappa}
\end{table}

\subsection{Feature Extraction}
Feature extraction is a crucial step in the modeling process, and it can help improve the model's overall performance by having more dimensions to factor in for the identification of the correct tags \cite{guyon2003introduction}. In this study, we extracted the following features covering word and sentence-based variables as listed below: 

\begin{enumerate}
    \item Boolean flags if the first letter of a target word is capitalized, all in uppercase or a digit.
    \item The character bigram and trigram of a target word.
    \item Whether a target word is at the beginning or end of the sentence (\texttt{BOS} or \texttt{EOS}).
    \item The two words to the left and the right of the target word.
    \item The top word clusters from an external embedding file for the target language.
\end{enumerate}


For the clustering component, we used a Cebuano corpus composed of Internet texts through the \textsc{cebtenten} corpus from Sketch Engine\footnote{\url{https://www.sketchengine.eu/cebtenten-cebuano-corpus/}}. 

\section{Modelling}
\begin{figure*}[!htbp]
    \centering
    \includegraphics[height=4.5cm]{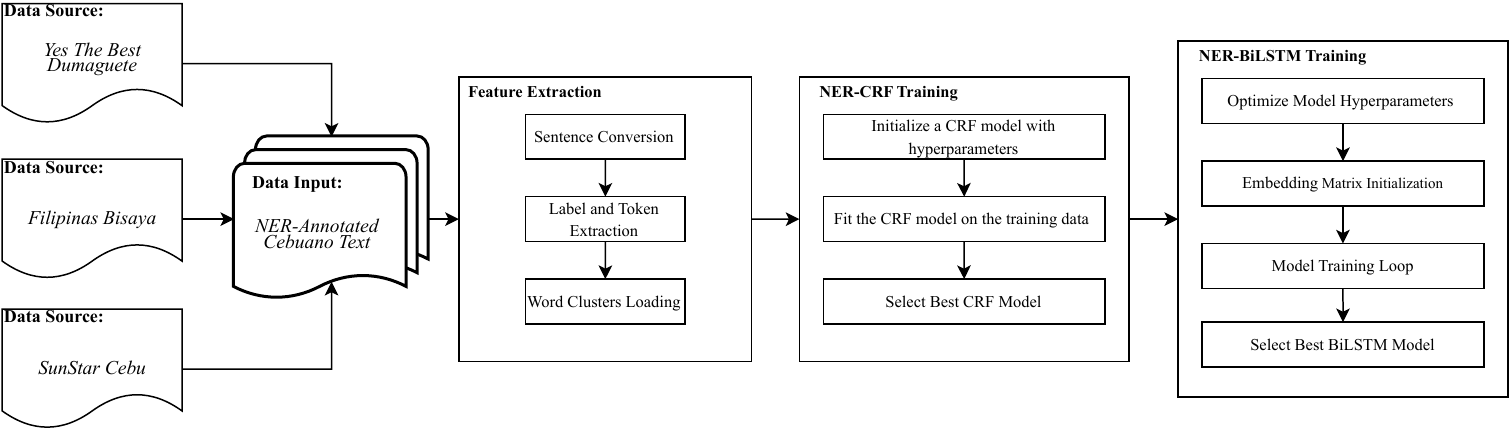}
    \caption{Overall methodology of developing \textsc{CebuaNER} using annotated news datasets in Cebuano with machine learning models CRF and BiLSTM.}
    \label{fig:methodology}
\end{figure*}

This section presents the modeling process that we used to develop our Cebuano NER system. To compare performance, we adopt two different techniques, \textbf{Conditional Random Field} (CRF) and \textbf{Bidirectional Long Short Term Memory} (BiLSTM) model. We use the package \texttt{sklearn-crfsuite} in Scikit-Learn \cite{pedregosa2011scikit} and \texttt{PyTorch} \cite{paszke2019pytorch} for the implementation of the training algorithms. We show a visual guide of the overall methodology of the study in Figure~\ref{fig:methodology}.


\subsection{Conditional Random Fields}
For the first modelling approach, we adopt one of the most common statistical methods for NER which is the Conditional Random Fields \cite{lafferty2001conditional}. CRFs are undirected graphical models that capture label conditional dependencies, making them ideal for applications such as part-of-speech tagging and named entity recognition \cite{ebona2013named,alfonso2013named,cotterell2017low}. Their ability to capture the relationships among adjacent words in a sentence is particularly valuable for NER since named entities often exhibit specific patterns in the context of the surrounding words \cite{wallach2004conditional}.

CRF architecture entails encoding the conditional probability distribution $P(y|x)$ over label sequences $Y$ given observation sequences $x$, enabling for quick and accurate sequence labeling without imposing unnecessary independence assumptions \cite{wallach2004conditional}. We show the main computation below where $\beta_t$ corresponds to the weight, $(y_t, y_{t-1}, x_t)$ for the feature, and ${Z}$ for the normalizing factor:

\begin{equation}
P(y|x) = \frac{1}{Z} \prod_{t=1}^T \beta_t(y_t, y_{t-1}, x_t)
\end{equation}


\subsection{Bidirectional Long Short-Term Memory}
For our second modelling approach, we advance to a neural network algorithm direction using Bidirectional Long Short-Term Memory or BiLSTM \cite{schuster1997bidirectional}. BiLSTM is a type of recurrent neural network (RNN) that has the ability to process sequential data in both forward and backward direction. It is also commonly used in tasks that involve sequence labeling such as NER with substantially larger datasets, in addition to being able to capture contextual information from both forward and backward direction of words in a sentence \cite{chiu-nichols-2016-named,reimers2017reporting,panchendrarajan2018bidirectional,zukov-gregoric-etal-2018-named}.


\section{Results}
In this section, we describe the outcome of training both the CRF and BiLSTM models using our newly-collected and annotation Cebuano NER dataset. Similar to previous works \cite{mayhew-etal-2019-named}, we omit the analysis with the OTH (other) tag as this usually serves as a miscellaneous label for more advanced tags in future annotations. 

\begin{table}[!htbp]
\small
\centering
\begin{tabular}{lcccc} 
 \toprule
 \textbf{Tagset} & \textbf{Precision} & \textbf{Recall} & \textbf{F1} & \textbf{Support} \\ 
 \midrule
 B-PER & 0.859 & 0.895 & 0.877 & 524  \\ 
 I-PER & 0.852 & 0.917 & 0.883 & 264  \\ 
 B-ORG & 0.825 & 0.558 & 0.665 & 312  \\ 
 I-ORG & 0.835 & 0.736 & 0.782 & 420  \\ 
 B-LOC & 0.854 & 0.731 & 0.788 & 383  \\ 
 I-LOC & 0.851 & 0.670 & 0.750 & 273  \\ 
 \bottomrule
\end{tabular}
\caption{Performance of the trained and \underline{un-optimized} CRF model for Cebuano NER}
\label{tab:crf_results_unop}
\end{table}

\begin{table}[!htbp]
\centering
\small
\begin{tabular}{lcccc} 
 \toprule
  \textbf{Tagset} & \textbf{Precision} & \textbf{Recall} & \textbf{F1} & \textbf{Support} \\ 
 \midrule
  B-PER & 0.881 & 0.918 & 0.899 & 524  \\ 
  I-PER & 0.875 & 0.932 & 0.903 & 264  \\ 
  B-ORG & 0.879 & 0.651 & 0.748 & 312  \\ 
  I-ORG & 0.860 & 0.729 & 0.789 & 420  \\ 
  B-LOC & 0.887 & 0.799 & 0.841 & 383  \\ 
  I-LOC & 0.833 & 0.733 & 0.780 & 273  \\ 
 \bottomrule
\end{tabular}
\caption{Performance of the trained and \underline{optimized} CRF model for Cebuano NER.}
\label{tab:crf_results}
\end{table}

For the CRF model, we first experimented with a standard optimization algorithm with LBFGS \cite{liu1989limited} that we ran for 100 iterations. Subsequently, a combination of L1 and L2 regularizations were used on the model to search for the optimal hyperparameters through a randomized search algorithm that we also ran for the same number of iterations to prevent overfitting. Upon evaluation of the resulting hyperparameters, we obtained an overall mean cross-validation F1 score of 0.901, 0.768, and 0.811 as calculated in Table~\ref{tab:crf_results} per tagset of PER, ORG, and LOC, respectively. We also note an overall improvement in performance from the initial evaluation from the unoptimized CRF model by about 2\%, 4\%, and 4.2\% per tagset of PER, ORG, and LOC, respectively.

\begin{table}[!htbp]
\centering
\small
\begin{tabular}{lcccc} 
 \toprule
 \textbf{Tagset} & \textbf{Precision} & \textbf{Recall} & \textbf{F1} & \textbf{Support} \\ 
 \midrule
  B-PER & 0.85 & 0.89 & 0.87 & 524  \\ 
 I-PER & 0.84 & 0.88 & 0.86 & 264  \\ 
 B-ORG & 0.78 & 0.36 & 0.49 & 312  \\ 
 I-ORG & 0.81 & 0.76 & 0.79 & 420  \\ 
 B-LOC & 0.85 & 0.69 & 0.76 & 383  \\ 
 I-LOC & 0.79 & 0.61 & 0.69 & 273  \\ 
 \bottomrule
\end{tabular}
\caption{Performance of the trained and \underline{optimized} BiLSTM model for Cebuano NER.}
\label{tab:bilstm_results}
\end{table}

Table~\ref{tab:bilstm_results} shows the results of model training for BiLSTM. The mean averages performance of the model for F1 score are 0.865, 0.640, and 0.725 per tagset of PER, ORG, and LOC, respectively. We observe that there is a close resemblance with the performance of the un-optimized CRF model in Table~\ref{tab:crf_results_unop}. We infer that this relatively lower performance can be attributed to the size of the data used. Specifically, this can be seen with the reduced performance in the F1 score, especially with identifying organization and location entities. Likewise, while CRFs are seen as the more traditional approach to work regarding NER, the use of BiLSTM may be more practical if the number of training data is higher than what we used. Despite our models being the new baseline due to having the highest training data used for Cebuano, future research works incorporating more annotated data should see an improvement across all performance metrics.


\section{Discussion}
In this section, we provide an in-depth discussion of insights obtained from the performances of the trained models, including error analysis and potential for crosslingual application.

\subsection{Error Analysis}
\begin{figure}[!htbp]
    \centering
    \includegraphics[width=1\linewidth]{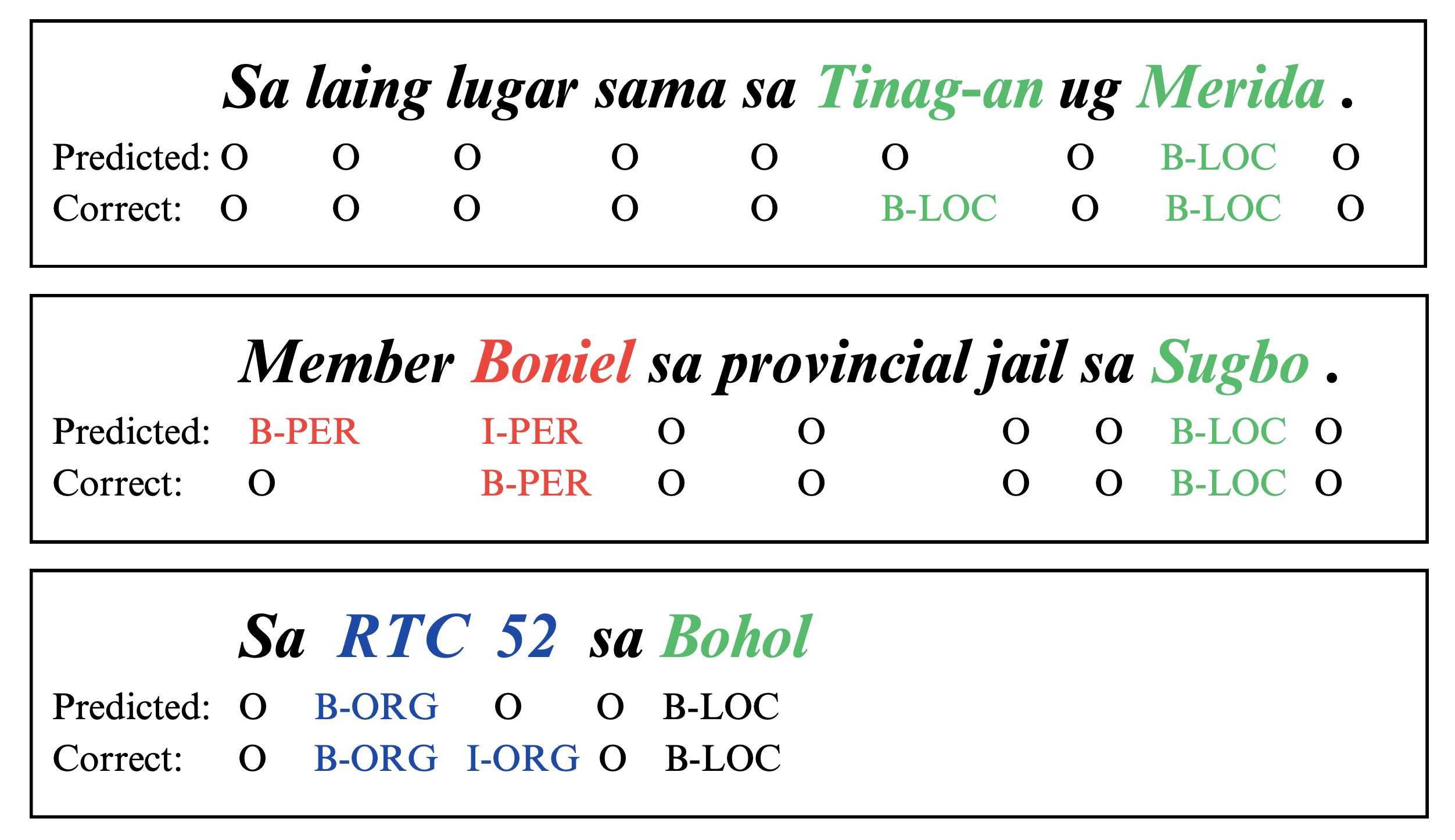}
    \caption{Cebuano sentences with misclassified annotations}
    \label{fig:miclassification}
\end{figure}

Where previous studies produced pronounced error rates when it came to identifying certain entities, such as in the works of \citet{alfonso2013named} and \citet{ebona2013named}, our best CRF model gives a more consistent performance that was specially fitted to the Cebuano language. However, several instances of misclassified tag predictions still occur, as shown with a few examples in Figure~\ref{fig:miclassification}. Within this subset, it was observed that certain named entities had been skipped by the model while other non-entity words were mistakenly labeled as qualified entities. For this, we further recommend further refinement of annotations, particularly with spans or entities longer than one word, in order to enhance the precision and efficiency of the NER model for this specific language.

\subsection{On Crosslingual Performance with Tagalog}
The crosslingual capability of NLP models, particular NER, is tested when a trained model using one language, in this case Cebuano, performs comparably well when tested on an unseen data in another language. This has been one of the features of NER systems that have been focused by previous works \cite{cotterell2017low,xie-etal-2018-neural,zhou-etal-2022-conner}. Although our goal for \textsc{CebuaNER} is to become a baseline model primarily for the Cebuano language, we still performed an initial crosslingual experiment show its potential to researchers interested in improving the model in the future. For this set, we used the best performing model which uses the CRF algorithm and a corrected version Tagalog dataset from the WikiANN data \cite{pan-etal-2017-cross} in the \texttt{calamanCy} library\footnote{\url{https://github.com/ljvmiranda921/calamanCy/tree/master}}. The Tagalog dataset contains 782 annotated documents with the same entity tag list of Person (PER), Organization (ORG), and Location (LOC). 

Table~\ref{tab:crosslingual_result} shows the performance of \textsc{CebuaNER} in a crosslingual setup with a Tagalog dataset. The mean averages in terms of F1 score are 0.713, 0.395, and 0.589 for entity tag list PER, ORG, and LOC respectively. While these are substantially lower overall compared to the previous CRF and BiLSTM models trained with purely Cebuano data in Tables~\ref{tab:crf_results} and ~\ref{tab:bilstm_results}, we see potential as recognition performance for identifying person and location names do not deviate too far. We also posit that Tagalog and Cebuano being members of the same language family subtree as seen in Figure~\ref{fig:cebuano_tree} also contribute to the two languages having overlapping linguistic intricacies such as grammar and word use \cite{imperial-etal-2022-baseline,imperial-kochmar-2023-automatic}.

\begin{table}[!htbp]
\centering
\small
\begin{tabular}{lcccc} 
 \toprule
 \textbf{Tagset} & \textbf{Precision} & \textbf{Recall} & \textbf{F1} & \textbf{Support} \\ 
 \midrule
 B-PER & 0.761 & 0.615 & 0.680 & 833  \\ 
 I-PER & 0.790 & 0.705 & 0.745 & 549  \\ 
 B-ORG & 0.386 & 0.303 & 0.340 & 363  \\ 
 I-ORG & 0.315 & 0.791 & 0.451 & 383  \\ 
 B-LOC & 0.766 & 0.436 & 0.556 & 383  \\ 
 I-LOC & 0.651 & 0.595 & 0.622 & 232  \\ 
 \bottomrule
\end{tabular}
\caption{Crosslingual experiment of the CRF-based \textsc{CebuaNER} model applied to a Tagalog test dataset.}
\label{tab:crosslingual_result}
\end{table}

\section{Conclusion}
Research initiatives involving the creation of high-quality corpus, release of technical implementations through code, and full transparency of model training are crucial to level the impact of low-resource languages in NLP. Towards contributing to this call, we introduced \textsc{CebuaNER}, a new baseline model for named entity recognition in the Cebuano language. \textsc{CebuaNER}'s main advantage from previous works is the use of a significantly larger gold-standard data from recent news articles to train models via CRF and BiLSTM, paired with empirical evidence of potential in a crosslingual application with Tagalog. In terms of performance, the best model for \textsc{CebuaNER} surpassed the mean standard threshold of 0.70 for precision, recall, and F1 across all entity tag list. We foresee that the public release of the trained models and annotated the dataset used will have substantial impact in the Philippine NLP landscape. Future works include improvements in span selection of the model to capture entities greater than one word as well as application to more complex neural network architectures if paired with an even higher data count.

\section*{Acknowledgment}
All datasets collected for this study are publicly available and are used for non-commercial research purposes. We acknowledge the sources of the Cebuano news articles being Yes the Best, Filipinas Bisaya, and Sunstar Cebu. This study is funded by the Philippine Commission on Higher Education (CHED) Leading the Advancement of Knowledge in Agriculture and Science (LAKAS) Project No. 2021-007, eParticipation 2.1: Harnessing Natural Language Processing (NLP) for Community Participation.

\bibliography{anthology,references}

\begin{thebibliography}{46}
\expandafter\ifx\csname natexlab\endcsname\relax\def\natexlab#1{#1}\fi

\bibitem[{Adelani et~al.(2022)Adelani, Neubig, Ruder, Rijhwani, Beukman,
  Palen-Michel, Lignos, Alabi, Muhammad, Nabende, Dione, Bukula, Mabuya,
  Dossou, Sibanda, Buzaaba, Mukiibi, Kalipe, Mbaye, Taylor, Kabore, Emezue,
  Aremu, Ogayo, Gitau, Munkoh-Buabeng, Memdjokam~Koagne, Tapo, Macucwa,
  Marivate, Elvis, Gwadabe, Adewumi, Ahia, Nakatumba-Nabende, Mokono, Ezeani,
  Chukwuneke, Oluwaseun~Adeyemi, Hacheme, Abdulmumin, Ogundepo, Yousuf, Moteu,
  and Klakow}]{adelani-etal-2022-masakhaner}
David Adelani, Graham Neubig, Sebastian Ruder, Shruti Rijhwani, Michael
  Beukman, Chester Palen-Michel, Constantine Lignos, Jesujoba Alabi,
  Shamsuddeen Muhammad, Peter Nabende, Cheikh M.~Bamba Dione, Andiswa Bukula,
  Rooweither Mabuya, Bonaventure F.~P. Dossou, Blessing Sibanda, Happy Buzaaba,
  Jonathan Mukiibi, Godson Kalipe, Derguene Mbaye, Amelia Taylor, Fatoumata
  Kabore, Chris~Chinenye Emezue, Anuoluwapo Aremu, Perez Ogayo, Catherine
  Gitau, Edwin Munkoh-Buabeng, Victoire Memdjokam~Koagne, Allahsera~Auguste
  Tapo, Tebogo Macucwa, Vukosi Marivate, Mboning~Tchiaze Elvis, Tajuddeen
  Gwadabe, Tosin Adewumi, Orevaoghene Ahia, Joyce Nakatumba-Nabende, Neo~Lerato
  Mokono, Ignatius Ezeani, Chiamaka Chukwuneke, Mofetoluwa Oluwaseun~Adeyemi,
  Gilles~Quentin Hacheme, Idris Abdulmumin, Odunayo Ogundepo, Oreen Yousuf,
  Tatiana Moteu, and Dietrich Klakow. 2022.
\newblock \href {https://aclanthology.org/2022.emnlp-main.298} {{M}asakha{NER}
  2.0: {A}frica-centric transfer learning for named entity recognition}.
\newblock In \emph{Proceedings of the 2022 Conference on Empirical Methods in
  Natural Language Processing}, pages 4488--4508, Abu Dhabi, United Arab
  Emirates. Association for Computational Linguistics.

\bibitem[{Adelani et~al.(2021)Adelani, Abbott, Neubig, D{'}souza, Kreutzer,
  Lignos, Palen-Michel, Buzaaba, Rijhwani, Ruder, Mayhew, Azime, Muhammad,
  Emezue, Nakatumba-Nabende, Ogayo, Anuoluwapo, Gitau, Mbaye, Alabi, Yimam,
  Gwadabe, Ezeani, Niyongabo, Mukiibi, Otiende, Orife, David, Ngom, Adewumi,
  Rayson, Adeyemi, Muriuki, Anebi, Chukwuneke, Odu, Wairagala, Oyerinde, Siro,
  Bateesa, Oloyede, Wambui, Akinode, Nabagereka, Katusiime, Awokoya, MBOUP,
  Gebreyohannes, Tilaye, Nwaike, Wolde, Faye, Sibanda, Ahia, Dossou, Ogueji,
  DIOP, Diallo, Akinfaderin, Marengereke, and
  Osei}]{adelani-etal-2021-masakhaner}
David~Ifeoluwa Adelani, Jade Abbott, Graham Neubig, Daniel D{'}souza, Julia
  Kreutzer, Constantine Lignos, Chester Palen-Michel, Happy Buzaaba, Shruti
  Rijhwani, Sebastian Ruder, Stephen Mayhew, Israel~Abebe Azime, Shamsuddeen~H.
  Muhammad, Chris~Chinenye Emezue, Joyce Nakatumba-Nabende, Perez Ogayo, Aremu
  Anuoluwapo, Catherine Gitau, Derguene Mbaye, Jesujoba Alabi, Seid~Muhie
  Yimam, Tajuddeen~Rabiu Gwadabe, Ignatius Ezeani, Rubungo~Andre Niyongabo,
  Jonathan Mukiibi, Verrah Otiende, Iroro Orife, Davis David, Samba Ngom, Tosin
  Adewumi, Paul Rayson, Mofetoluwa Adeyemi, Gerald Muriuki, Emmanuel Anebi,
  Chiamaka Chukwuneke, Nkiruka Odu, Eric~Peter Wairagala, Samuel Oyerinde,
  Clemencia Siro, Tobius~Saul Bateesa, Temilola Oloyede, Yvonne Wambui, Victor
  Akinode, Deborah Nabagereka, Maurice Katusiime, Ayodele Awokoya, Mouhamadane
  MBOUP, Dibora Gebreyohannes, Henok Tilaye, Kelechi Nwaike, Degaga Wolde,
  Abdoulaye Faye, Blessing Sibanda, Orevaoghene Ahia, Bonaventure F.~P. Dossou,
  Kelechi Ogueji, Thierno~Ibrahima DIOP, Abdoulaye Diallo, Adewale Akinfaderin,
  Tendai Marengereke, and Salomey Osei. 2021.
\newblock \href {https://doi.org/10.1162/tacl_a_00416} {{M}asakha{NER}: Named
  entity recognition for {A}frican languages}.
\newblock \emph{Transactions of the Association for Computational Linguistics},
  9:1116--1131.

\bibitem[{Aji et~al.(2022)Aji, Winata, Koto, Cahyawijaya, Romadhony, Mahendra,
  Kurniawan, Moeljadi, Prasojo, Baldwin, Lau, and Ruder}]{aji-etal-2022-one}
Alham~Fikri Aji, Genta~Indra Winata, Fajri Koto, Samuel Cahyawijaya, Ade
  Romadhony, Rahmad Mahendra, Kemal Kurniawan, David Moeljadi, Radityo~Eko
  Prasojo, Timothy Baldwin, Jey~Han Lau, and Sebastian Ruder. 2022.
\newblock \href {https://doi.org/10.18653/v1/2022.acl-long.500} {One country,
  700+ languages: {NLP} challenges for underrepresented languages and dialects
  in {I}ndonesia}.
\newblock In \emph{Proceedings of the 60th Annual Meeting of the Association
  for Computational Linguistics (Volume 1: Long Papers)}, pages 7226--7249,
  Dublin, Ireland. Association for Computational Linguistics.

\bibitem[{Alfonso et~al.(2013)Alfonso, Domingo, Galope, Sagum, Villar, and
  Villegas}]{alfonso2013named}
Ana Patricia~T Alfonso, Illuminada Vivien~R Domingo, Mary Joy~F Galope, Ria~A
  Sagum, Rachelle~B Villar, and Jobert~T Villegas. 2013.
\newblock Named entity recognizer for filipino text using conditional random
  field.
\newblock \emph{International Journal of Future Computer and Communication},
  2(5):376.

\bibitem[{Balasuriya et~al.(2009)Balasuriya, Ringland, Nothman, Murphy, and
  Curran}]{balasuriya2009named}
Dominic Balasuriya, Nicky Ringland, Joel Nothman, Tara Murphy, and James~R
  Curran. 2009.
\newblock Named entity recognition in wikipedia.
\newblock In \emph{Proceedings of the 2009 workshop on the people’s web meets
  NLP: Collaboratively constructed semantic resources (People’s Web)}, pages
  10--18.

\bibitem[{Brandsen et~al.(2020)Brandsen, Verberne, Wansleeben, and
  Lambers}]{brandsen2020creating}
Alex Brandsen, Suzan Verberne, Milco Wansleeben, and Karsten Lambers. 2020.
\newblock Creating a dataset for named entity recognition in the archaeology
  domain.
\newblock In \emph{Proceedings of the Twelfth Language Resources and Evaluation
  Conference}, pages 4573--4577.

\bibitem[{Cai et~al.(2023)Cai, Huang, Jiang, Tan, Xie, and
  Tu}]{cai-etal-2023-graph}
Jiong Cai, Shen Huang, Yong Jiang, Zeqi Tan, Pengjun Xie, and Kewei Tu. 2023.
\newblock \href {https://aclanthology.org/2023.acl-short.11} {Graph propagation
  based data augmentation for named entity recognition}.
\newblock In \emph{Proceedings of the 61st Annual Meeting of the Association
  for Computational Linguistics (Volume 2: Short Papers)}, pages 110--118,
  Toronto, Canada. Association for Computational Linguistics.

\bibitem[{Chiu and Nichols(2016)}]{chiu-nichols-2016-named}
Jason~P.C. Chiu and Eric Nichols. 2016.
\newblock \href {https://doi.org/10.1162/tacl_a_00104} {Named entity
  recognition with bidirectional {LSTM}-{CNN}s}.
\newblock \emph{Transactions of the Association for Computational Linguistics},
  4:357--370.

\bibitem[{Cohen(1960)}]{cohen1960coefficient}
Jacob Cohen. 1960.
\newblock A coefficient of agreement for nominal scales.
\newblock \emph{Educational and psychological measurement}, 20(1):37--46.

\bibitem[{Cotterell and Duh(2017{\natexlab{a}})}]{cotterell-duh-2017-low}
Ryan Cotterell and Kevin Duh. 2017{\natexlab{a}}.
\newblock \href {https://aclanthology.org/I17-2016} {Low-resource named entity
  recognition with cross-lingual, character-level neural conditional random
  fields}.
\newblock In \emph{Proceedings of the Eighth International Joint Conference on
  Natural Language Processing (Volume 2: Short Papers)}, pages 91--96, Taipei,
  Taiwan. Asian Federation of Natural Language Processing.

\bibitem[{Cotterell and Duh(2017{\natexlab{b}})}]{cotterell2017low}
Ryan Cotterell and Kevin Duh. 2017{\natexlab{b}}.
\newblock Low-resource named entity recognition with cross-lingual,
  character-level neural conditional random fields.
\newblock In \emph{Proceedings of the Eighth International Joint Conference on
  Natural Language Processing (Volume 2: Short Papers)}, pages 91--96.

\bibitem[{Cruz et~al.(2018)Cruz, Montalla, Manansala, Rodriguez, Octaviano, and
  Fabito}]{cruz2018named}
Bern Maris~Dela Cruz, Cyril Montalla, Allysa Manansala, Ramon Rodriguez,
  Manolito Octaviano, and Bernie~S Fabito. 2018.
\newblock Named-entity recognition for disaster related filipino news articles.
\newblock In \emph{TENCON 2018-2018 IEEE Region 10 Conference}, pages
  1633--1636. IEEE.

\bibitem[{Ebo{\~n}a et~al.(2013)Ebo{\~n}a, Llorca~Jr, Perez, Roldan, Domingo,
  and Sagum}]{ebona2013named}
Karen Mae~L Ebo{\~n}a, Orlando~S Llorca~Jr, Genrev~P Perez, Jhustine~M Roldan,
  Iluminda Vivien~R Domingo, and Ria~A Sagum. 2013.
\newblock Named-entity recognizer (ner) for filipino novel excerpts using
  maximum entropy approach.
\newblock \emph{Journal of Industrial and Intelligent Information Vol}, 1(1).

\bibitem[{Gharagozlou et~al.(2023)Gharagozlou, Mohammadzadeh, Bastanfard, and
  Ghidary}]{gharagozlou2023semantic}
Hamid Gharagozlou, Javad Mohammadzadeh, Azam Bastanfard, and Saeed~Shiry
  Ghidary. 2023.
\newblock Semantic relation extraction: A review of approaches, datasets, and
  evaluation methods.
\newblock \emph{ACM Transactions on Asian and Low-Resource Language Information
  Processing}.

\bibitem[{Gonzales et~al.(2022)Gonzales, Gustilo, Nituda, and
  Adlaon}]{gonzales2022developing}
Joshua Andre~Huertas Gonzales, J-Adrielle~Enriquez Gustilo, Glenn
  Michael~Vequilla Nituda, and Kristine Mae~Monteza Adlaon. 2022.
\newblock Developing a hybrid neural network for part-of-speech tagging and
  named entity recognition.
\newblock In \emph{Proceedings of the 2022 5th Artificial Intelligence and
  Cloud Computing Conference}, pages 7--13.

\bibitem[{Guyon and Elisseeff(2003)}]{guyon2003introduction}
Isabelle Guyon and Andr{\'e} Elisseeff. 2003.
\newblock An introduction to variable and feature selection.
\newblock \emph{Journal of machine learning research}, 3(Mar):1157--1182.

\bibitem[{ImaniGooghari et~al.(2023)ImaniGooghari, Lin, Kargaran, Severini,
  Jalili~Sabet, Kassner, Ma, Schmid, Martins, Yvon, and
  Sch{\"u}tze}]{imanigooghari-etal-2023-glot500}
Ayyoob ImaniGooghari, Peiqin Lin, Amir~Hossein Kargaran, Silvia Severini,
  Masoud Jalili~Sabet, Nora Kassner, Chunlan Ma, Helmut Schmid, Andr{\'e}
  Martins, Fran{\c{c}}ois Yvon, and Hinrich Sch{\"u}tze. 2023.
\newblock \href {https://aclanthology.org/2023.acl-long.61} {Glot500: Scaling
  multilingual corpora and language models to 500 languages}.
\newblock In \emph{Proceedings of the 61st Annual Meeting of the Association
  for Computational Linguistics (Volume 1: Long Papers)}, pages 1082--1117,
  Toronto, Canada. Association for Computational Linguistics.

\bibitem[{Imperial and Kochmar(2023)}]{imperial-kochmar-2023-automatic}
Joseph~Marvin Imperial and Ekaterina Kochmar. 2023.
\newblock \href {https://aclanthology.org/2023.findings-acl.331} {Automatic
  readability assessment for closely related languages}.
\newblock In \emph{Findings of the Association for Computational Linguistics:
  ACL 2023}, pages 5371--5386, Toronto, Canada. Association for Computational
  Linguistics.

\bibitem[{Imperial et~al.(2022)Imperial, Reyes, Ibanez, Sapinit, and
  Hussien}]{imperial-etal-2022-baseline}
Joseph~Marvin Imperial, Lloyd Lois~Antonie Reyes, Michael~Antonio Ibanez, Ranz
  Sapinit, and Mohammed Hussien. 2022.
\newblock \href {https://doi.org/10.18653/v1/2022.bea-1.5} {A baseline
  readability model for {C}ebuano}.
\newblock In \emph{Proceedings of the 17th Workshop on Innovative Use of NLP
  for Building Educational Applications (BEA 2022)}, pages 27--32, Seattle,
  Washington. Association for Computational Linguistics.

\bibitem[{Jarrar et~al.(2022)Jarrar, Khalilia, and
  Ghanem}]{jarrar-etal-2022-wojood}
Mustafa Jarrar, Mohammed Khalilia, and Sana Ghanem. 2022.
\newblock \href {https://aclanthology.org/2022.lrec-1.387} {Wojood: Nested
  {A}rabic named entity corpus and recognition using {BERT}}.
\newblock In \emph{Proceedings of the Thirteenth Language Resources and
  Evaluation Conference}, pages 3626--3636, Marseille, France. European
  Language Resources Association.

\bibitem[{Lafferty et~al.(2001)Lafferty, McCallum, and
  Pereira}]{lafferty2001conditional}
John~D Lafferty, Andrew McCallum, and Fernando~CN Pereira. 2001.
\newblock Conditional random fields: Probabilistic models for segmenting and
  labeling sequence data.
\newblock In \emph{Proceedings of the Eighteenth International Conference on
  Machine Learning}, pages 282--289.

\bibitem[{Liu and Nocedal(1989)}]{liu1989limited}
Dong~C Liu and Jorge Nocedal. 1989.
\newblock On the limited memory bfgs method for large scale optimization.
\newblock \emph{Mathematical programming}, 45(1-3):503--528.

\bibitem[{Liu et~al.(2018)Liu, Shang, Ren, Xu, Gui, Peng, and
  Han}]{liu2018empower}
Liyuan Liu, Jingbo Shang, Xiang Ren, Frank Xu, Huan Gui, Jian Peng, and Jiawei
  Han. 2018.
\newblock Empower sequence labeling with task-aware neural language model.
\newblock In \emph{Proceedings of the AAAI Conference on Artificial
  Intelligence}, volume~32.

\bibitem[{Lorica and Nathan(2021)}]{loricanathan}
Ben Lorica and Paco Nathan. 2021.
\newblock 2021 nlp survey report.

\bibitem[{Mayhew et~al.(2019)Mayhew, Chaturvedi, Tsai, and
  Roth}]{mayhew-etal-2019-named}
Stephen Mayhew, Snigdha Chaturvedi, Chen-Tse Tsai, and Dan Roth. 2019.
\newblock \href {https://doi.org/10.18653/v1/K19-1060} {Named entity
  recognition with partially annotated training data}.
\newblock In \emph{Proceedings of the 23rd Conference on Computational Natural
  Language Learning (CoNLL)}, pages 645--655, Hong Kong, China. Association for
  Computational Linguistics.

\bibitem[{Mayhew and Roth(2018)}]{mayhew-roth-2018-talen}
Stephen Mayhew and Dan Roth. 2018.
\newblock \href {https://doi.org/10.18653/v1/P18-4014} {{TALEN}: Tool for
  annotation of low-resource {EN}tities}.
\newblock In \emph{Proceedings of {ACL} 2018, System Demonstrations}, pages
  80--86, Melbourne, Australia. Association for Computational Linguistics.

\bibitem[{Maynard et~al.(2003)Maynard, Tablan, and Cunningham}]{maynard2003ne}
Diana Maynard, Valentin Tablan, and Hamish Cunningham. 2003.
\newblock Ne recognition without training data on a language you don’t speak.
\newblock In \emph{Proceedings of the ACL 2003 workshop on multilingual and
  mixed-language named entity recognition}, pages 33--40.

\bibitem[{McFarland(2008)}]{mcfarland2008linguistic}
Curtis~D McFarland. 2008.
\newblock Linguistic diversity and english in the philippines.
\newblock \emph{Philippine English: Linguistic and literary perspectives},
  1:131.

\bibitem[{Metila et~al.(2016)Metila, Pradilla, and
  Williams}]{metila2016challenge}
Romylyn~A Metila, Lea Angela~S Pradilla, and Alan~B Williams. 2016.
\newblock The challenge of implementing mother tongue education in
  linguistically diverse contexts: The case of the philippines.
\newblock \emph{The Asia-Pacific Education Researcher}, 25:781--789.

\bibitem[{Pan et~al.(2017)Pan, Zhang, May, Nothman, Knight, and
  Ji}]{pan-etal-2017-cross}
Xiaoman Pan, Boliang Zhang, Jonathan May, Joel Nothman, Kevin Knight, and Heng
  Ji. 2017.
\newblock \href {https://doi.org/10.18653/v1/P17-1178} {Cross-lingual name
  tagging and linking for 282 languages}.
\newblock In \emph{Proceedings of the 55th Annual Meeting of the Association
  for Computational Linguistics (Volume 1: Long Papers)}, pages 1946--1958,
  Vancouver, Canada. Association for Computational Linguistics.

\bibitem[{Panchendrarajan and
  Amaresan(2018)}]{panchendrarajan2018bidirectional}
Rrubaa Panchendrarajan and Aravindh Amaresan. 2018.
\newblock Bidirectional lstm-crf for named entity recognition.
\newblock In \emph{Proceedings of the 32nd Pacific Asia Conference on Language,
  Information and Computation}.

\bibitem[{Pant et~al.(2023)Pant, Sharma, and Kundu}]{pant2023named}
Vinay~Kumar Pant, Rupak Sharma, and Shakti Kundu. 2023.
\newblock Named entity recognition of kumauni language using machine learning
  (ml).

\bibitem[{Paszke et~al.(2019)Paszke, Gross, Massa, Lerer, Bradbury, Chanan,
  Killeen, Lin, Gimelshein, Antiga et~al.}]{paszke2019pytorch}
Adam Paszke, Sam Gross, Francisco Massa, Adam Lerer, James Bradbury, Gregory
  Chanan, Trevor Killeen, Zeming Lin, Natalia Gimelshein, Luca Antiga, et~al.
  2019.
\newblock Pytorch: An imperative style, high-performance deep learning library.
\newblock \emph{Advances in neural information processing systems}, 32.

\bibitem[{Pedregosa et~al.(2011)Pedregosa, Varoquaux, Gramfort, Michel,
  Thirion, Grisel, Blondel, Prettenhofer, Weiss, Dubourg
  et~al.}]{pedregosa2011scikit}
Fabian Pedregosa, Ga{\"e}l Varoquaux, Alexandre Gramfort, Vincent Michel,
  Bertrand Thirion, Olivier Grisel, Mathieu Blondel, Peter Prettenhofer, Ron
  Weiss, Vincent Dubourg, et~al. 2011.
\newblock Scikit-learn: Machine learning in python.
\newblock \emph{the Journal of machine Learning research}, 12:2825--2830.

\bibitem[{Reimers and Gurevych(2017)}]{reimers2017reporting}
Nils Reimers and Iryna Gurevych. 2017.
\newblock Reporting score distributions makes a difference: Performance study
  of lstm-networks for sequence tagging.
\newblock \emph{arXiv preprint arXiv:1707.09861}.

\bibitem[{Rijhwani et~al.(2020)Rijhwani, Zhou, Neubig, and
  Carbonell}]{rijhwani-etal-2020-soft}
Shruti Rijhwani, Shuyan Zhou, Graham Neubig, and Jaime Carbonell. 2020.
\newblock \href {https://doi.org/10.18653/v1/2020.acl-main.722} {Soft
  gazetteers for low-resource named entity recognition}.
\newblock In \emph{Proceedings of the 58th Annual Meeting of the Association
  for Computational Linguistics}, pages 8118--8123, Online. Association for
  Computational Linguistics.

\bibitem[{Schuster and Paliwal(1997)}]{schuster1997bidirectional}
Mike Schuster and Kuldip~K Paliwal. 1997.
\newblock Bidirectional recurrent neural networks.
\newblock \emph{IEEE transactions on Signal Processing}, 45(11):2673--2681.

\bibitem[{Simons et~al.(2022)Simons, Thomas, and
  White}]{simons-etal-2022-assessing}
Gary~F. Simons, Abbey L.~L. Thomas, and Chad K.~K. White. 2022.
\newblock \href {https://aclanthology.org/2022.coling-1.379} {Assessing digital
  language support on a global scale}.
\newblock In \emph{Proceedings of the 29th International Conference on
  Computational Linguistics}, pages 4299--4305, Gyeongju, Republic of Korea.
  International Committee on Computational Linguistics.

\bibitem[{Vajjala and
  Balasubramaniam(2022)}]{vajjala-balasubramaniam-2022-really}
Sowmya Vajjala and Ramya Balasubramaniam. 2022.
\newblock \href {https://aclanthology.org/2022.lrec-1.643} {What do we really
  know about state of the art {NER}?}
\newblock In \emph{Proceedings of the Thirteenth Language Resources and
  Evaluation Conference}, pages 5983--5993, Marseille, France. European
  Language Resources Association.

\bibitem[{Wallach(2004)}]{wallach2004conditional}
Hanna~M Wallach. 2004.
\newblock Conditional random fields: An introduction.
\newblock \emph{Technical Reports (CIS)}, page~22.

\bibitem[{Wilie et~al.(2020)Wilie, Vincentio, Winata, Cahyawijaya, Li, Lim,
  Soleman, Mahendra, Fung, Bahar, and Purwarianti}]{wilie-etal-2020-indonlu}
Bryan Wilie, Karissa Vincentio, Genta~Indra Winata, Samuel Cahyawijaya,
  Xiaohong Li, Zhi~Yuan Lim, Sidik Soleman, Rahmad Mahendra, Pascale Fung,
  Syafri Bahar, and Ayu Purwarianti. 2020.
\newblock \href {https://aclanthology.org/2020.aacl-main.85} {{I}ndo{NLU}:
  Benchmark and resources for evaluating {I}ndonesian natural language
  understanding}.
\newblock In \emph{Proceedings of the 1st Conference of the Asia-Pacific
  Chapter of the Association for Computational Linguistics and the 10th
  International Joint Conference on Natural Language Processing}, pages
  843--857, Suzhou, China. Association for Computational Linguistics.

\bibitem[{Xie et~al.(2018)Xie, Yang, Neubig, Smith, and
  Carbonell}]{xie-etal-2018-neural}
Jiateng Xie, Zhilin Yang, Graham Neubig, Noah~A. Smith, and Jaime Carbonell.
  2018.
\newblock \href {https://doi.org/10.18653/v1/D18-1034} {Neural cross-lingual
  named entity recognition with minimal resources}.
\newblock In \emph{Proceedings of the 2018 Conference on Empirical Methods in
  Natural Language Processing}, pages 369--379, Brussels, Belgium. Association
  for Computational Linguistics.

\bibitem[{Yaseen and Langer(2021)}]{yaseen-langer-2021-data}
Usama Yaseen and Stefan Langer. 2021.
\newblock \href {https://aclanthology.org/2021.icon-main.43} {Data augmentation
  for low-resource named entity recognition using backtranslation}.
\newblock In \emph{Proceedings of the 18th International Conference on Natural
  Language Processing (ICON)}, pages 352--358, National Institute of Technology
  Silchar, Silchar, India. NLP Association of India (NLPAI).

\bibitem[{Zhou et~al.(2019)Zhou, Zhang, Jin, Zhu, Fang, Goh, and
  Kwok}]{zhou-etal-2019-dual}
Joey~Tianyi Zhou, Hao Zhang, Di~Jin, Hongyuan Zhu, Meng Fang, Rick Siow~Mong
  Goh, and Kenneth Kwok. 2019.
\newblock \href {https://doi.org/10.18653/v1/P19-1336} {Dual adversarial neural
  transfer for low-resource named entity recognition}.
\newblock In \emph{Proceedings of the 57th Annual Meeting of the Association
  for Computational Linguistics}, pages 3461--3471, Florence, Italy.
  Association for Computational Linguistics.

\bibitem[{Zhou et~al.(2022)Zhou, Li, Bing, Cambria, Si, and
  Miao}]{zhou-etal-2022-conner}
Ran Zhou, Xin Li, Lidong Bing, Erik Cambria, Luo Si, and Chunyan Miao. 2022.
\newblock \href {https://aclanthology.org/2022.emnlp-main.577} {{C}on{NER}:
  Consistency training for cross-lingual named entity recognition}.
\newblock In \emph{Proceedings of the 2022 Conference on Empirical Methods in
  Natural Language Processing}, pages 8438--8449, Abu Dhabi, United Arab
  Emirates. Association for Computational Linguistics.

\bibitem[{{\v{Z}}ukov-Gregori{\v{c}} et~al.(2018){\v{Z}}ukov-Gregori{\v{c}},
  Bachrach, and Coope}]{zukov-gregoric-etal-2018-named}
Andrej {\v{Z}}ukov-Gregori{\v{c}}, Yoram Bachrach, and Sam Coope. 2018.
\newblock \href {https://doi.org/10.18653/v1/P18-2012} {Named entity
  recognition with parallel recurrent neural networks}.
\newblock In \emph{Proceedings of the 56th Annual Meeting of the Association
  for Computational Linguistics (Volume 2: Short Papers)}, pages 69--74,
  Melbourne, Australia. Association for Computational Linguistics.

\end{thebibliography}
\bibliographystyle{acl_natbib}




\end{document}